\documentclass{article}

\usepackage{microtype}
\usepackage{graphicx}
\usepackage{subfigure}
\usepackage{booktabs} 

\usepackage{hyperref}

\usepackage[accepted]{icml2024}

\usepackage{amsmath}
\usepackage{amssymb}
\usepackage{mathtools}
\usepackage{amsthm}
\usepackage{colortbl}

\usepackage[capitalize,noabbrev]{cleveref}

\theoremstyle{plain}

\theoremstyle{definition}

\theoremstyle{remark}

\usepackage[textsize=tiny]{todonotes}

\AtBeginDocument{%
  \providecommand\BibTeX{{%
    \normalfont B\kern-0.5em{\scshape i\kern-0.25em b}\kern-0.8em\TeX}}}

\def\numx#1e#2{{#1}\mathrm{e}{#2}}

\icmltitlerunning{}

\begin{document}
\twocolumn[
\icmltitle{\textsc{PokéLLMon}: A Human-Parity Agent for Pokémon Battles \\ with Large Language Models}

\newcommand{\weburl}{\url{https://poke-llm-on.github.io/}}
\newcommand{\sch}{Georgia Institute of Technology \\ Atlanta, GA 30332, United States \\ \texttt{\{sihaohu, thuang, ling.liu\}@gatech.edu} }

\begin{icmlauthorlist}
\icmlauthor{Sihao Hu, Tiansheng Huang, Ling Liu}{}
\end{icmlauthorlist}
\centering

\sch

\weburl

\icmlcorrespondingauthor{}{}

\icmlkeywords{Machine Learning, ICML}

\vskip 0.3in
]


\begin{abstract}

We introduce \textsc{PokéLLMon}, the first LLM-based agent that achieves human-parity performance in tactical battle games, as demonstrated in Pokémon battles. The design of \textsc{PokéLLMon} incorporates three key strategies: (i) In-context reinforcement learning that instantly consumes text-based feedback derived from battles to iteratively refine the policy; (ii) Knowledge-augmented generation that retrieves external knowledge to counteract hallucination and enables the agent to act timely and properly; (iii) Consistent action generation to mitigate the \textit{panic switching} phenomenon when the agent faces a powerful opponent and wants to elude
the battle. We show that online battles against human demonstrates \textsc{PokéLLMon}'s human-like battle strategies and just-in-time decision making, achieving 49\% of win rate in the Ladder competitions and 56\% of win rate in the invited battles. 
Our implementation and playable battle logs are available at: 
{\small\url{https://github.com/git-disl/PokeLLMon}}.

\end{abstract}

\section{Introduction}

Generative AI and Large Language Models (LLMs) have shown unprecedented success on NLP tasks~\cite{chatGPT,GPT3,xi2023rise,wang2023survey}. One of the forthcoming advancements will be to explore how LLMs can autonomously act in the physical world with extended generation space from text to action, representing a pivotal paradigm in the pursuit of Artificial General Intelligence~\cite{goertzel2007artificial,goertzel2014artificial}. 

Games are suitable test-beds to develop LLM-based agents~\cite{duan2022survey,batra2020rearrangement} to interact with the virtual environment in a way resembling human behavior. For example, Generative Agents~\cite{park2023generative} conducts a social experiments with LLMs assuming various roles in a ``The Sims"-like sandbox, where agents exhbit behavior and social interactions mirroring humans. In Mincraft, decision-making agents~\cite{wang2023voyager,wang2023describe,singh2023progprompt} are designed to explore the world and develop new skills for solving tasks and making tools.

\begin{figure}[tbp]
    \centering
    \includegraphics[width=6cm]{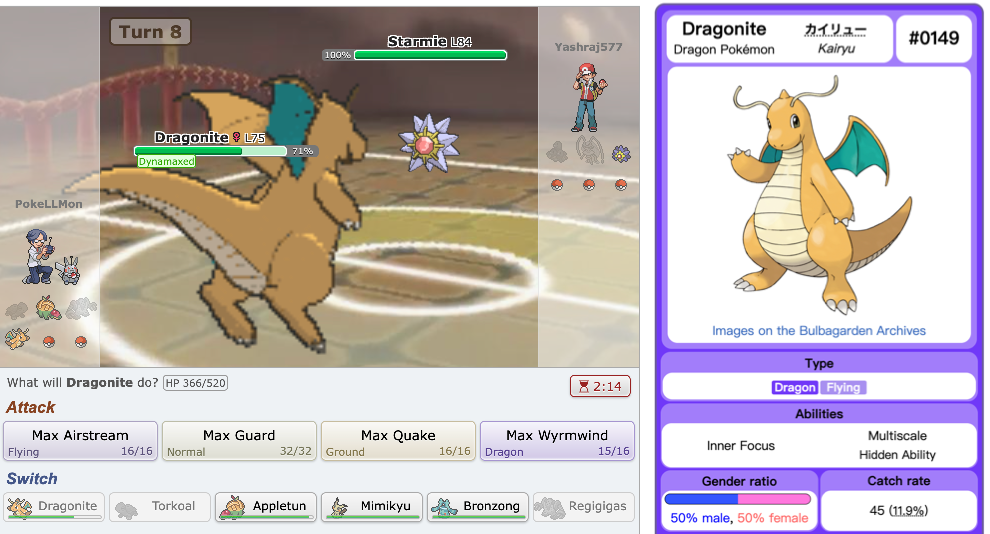}
    \vspace{-0.2cm}
    \caption{{\small At each turn, the player is requested to decide which action to perform, \textit{i.e.}, whether to let \textit{Dragonite} to take a move or switch to another Pokémon off the field.}}
    \vspace{0.3cm}
    \label{fig:pokemon_battle}
\end{figure}

Compared to existing games, tactical battle games~\cite{starcraftii} are better suited for benchmarking the game-playing ability of LLMs as the win rate can be directly measured and consistent opponents like AI or human players are always available. Pokémon battles, serving as a mechanism that evaluates the battle abilities of trainers in the well-known Pokémon games, offer several unique advantages as the first attempt for LLMs to play tactical battle games: 

(1) The state and action spaces are discrete and can be translated into text losslessly. \textbf{Figure}~\ref{fig:pokemon_battle} is an illustrative example for a Pokémon battle: At each turn, the player is requested to generate an action to perform given the current state of Pokémon from each side. The action space consists of four moves and five possible Pokémon to switch; (2) The turn-based format eliminates the demands of intensive gameplay, alleviating the stress on the inference time cost for LLMs, making performance hinges solely on the reasoning abilities of LLMs; (3) Despite its seemingly simple mechanism, Pokémon battle is strategic and complex: an experienced player takes various factors into consideration, including species/type/ability/stats/item/moves of all the Pokémon on and off the field. In a random battle, each Pokémon is randomly selected from a large candidate pool (more than 1,000) with distinct characteristics, demanding the players both the Pokémon knowledge and reasoning ability. 


\textbf{Scope and Contributions:} The scope of this paper is to develop an LLM-based agent that mimics the way a human player engages in Pokémon battles. The objective is to explore the key factors that make the LLM-based agent a good player and to examine its strengths and weaknesses in battles against human players.
%
%
To enable LLMs play game autonomously, we implement an environment that can parse and translate battle state into text description, and deliver generated action back to the server. By evaluating existing LLMs, we identify the presence of \textit{hallucination}, and the \textit{panic switching} phenomenon. 

\textit{Hallucination}: The agent can mistakenly send out Pokémon at a type disadvantage or persist in using ineffective moves against the opponent. As a result, the most advanced LLM, GPT-4, achieves a win rate of 26\% when playing against a heuristic bot, compared to 60\% win rate of human players. To combat hallucination, we introduce two strategies: (1) In-context reinforcement learning: We provide the agent with text-based feedback \textit{instantly} derived from the battle, serving as a new form of ``reward" to \textit{iteratively} refine the action generation policy without training; (2) Knowledge-augmented generation: We equip the agent with Pokédex, an encyclopaedia in Pokémon games that provides external knowledge like type advantage relationship or move/ability descriptions, simulating a human player searching for the information of unfamiliar Pokémon.

\textit{Panic switching}: We discover that when the agent encounters a powerful Pokémon, it tends to panic and generates inconsistent actions like  switching different Pokémon in consecutive turns to elude the battle, a phenomenon that is especially pronounced with Chain-of-Thought~\cite{CoT} reasoning. Consistent action generation alleviates the issue by voting out the most consistent action without overthinking. This observation mirrors human behavior, where in stressful situations, overthinking and exaggerating difficulties can lead to panic and impede acting.

Online battles demonstrate {\small \textsc{PokéLLMon}}'s human-competitive battle abilities: it achieves a 49\% win rate in the Ladder competitions and a 56\% win rate in the invited battles. Furthermore, we reveal its vulnerabilities to human players' attrition strategies and deceptive tricks.

In summary, this paper makes four original contributions: 
%
\begin{itemize}
   \item We implement and release an environment that enables LLMs to autonomously play Pokémon battles.
   \item We propose in-context reinforcement learning to instantly and iteratively refine the policy, and knowledge-augmented generation to combat hallucination.  
 \item We discover that the agent with chain-of-thought experiences panic when facing powerful opponents, and consistent action generation can mitigate this issue.
  \item {\small \textsc{PokéLLMon}}, to the best of our knowledge, is the first LLM-based agent with human-parity performance in tactical battle games.
\end{itemize}


\section{LLMs as Game Players}


\textbf{Communicative games:} Communicative games revolve around communication, deduction and sometimes deception between players. Existing studies show that LLMs demonstrate strategic behaviors in board games like Werewolf~\cite{werewolf}, Avalane~\cite{avalon}, WorldWar II~\cite{worldwarii} and Diplomacy~\cite{Cicero}.



\textbf{Open-ended games:} Open-ended games allow players to freely explore the game world and interact with others. Generative Agent~\cite{park2023generative} showcases that LLM-based agents exhibit behavior and social interactions mirroring human-like patterns. In MineCraft, Voyager~\cite{wang2023voyager} employs curriculum mechanism to explore the world and generates and executes code for solving tasks. DEPS~\cite{wang2023describe} proposes an approach of ``Describe, Explain, Plan and Select" to accomplish 70+ tasks. Planing-based frameworks like AutoGPT~\cite{Significant_Gravitas_AutoGPT} and MetaGPT~\cite{hong2023metagpt} can be adopted for the exploration task as well.

\textbf{Tactic battle games:} Among various game types, tactical battle games~\cite{akata2023playing, starcraftii} are particularly suitable for benchmarking LLMs' game-playing ability, as the win rate can be directly measured, and consistent opponents are always available. Recently, LLMs are employed to play StarCraft II~\cite{starcraftii} against the built-in AI with a text-based interface and a chain-of-summarization approach. In comparison, {\small \textsc{PokéLLMon}} has several advantages: (1) Translating Pokémon battle state into text is lossless; (2) Turn-based format eliminates real-time stress given the inference time cost of LLMs; (3) Battling against disciplined human players elevates the difficulty to a new height.


\section{Background}

\subsection{Pokémon}

\begin{figure}[tbp]
    \centering
    \includegraphics[width=7.0cm]{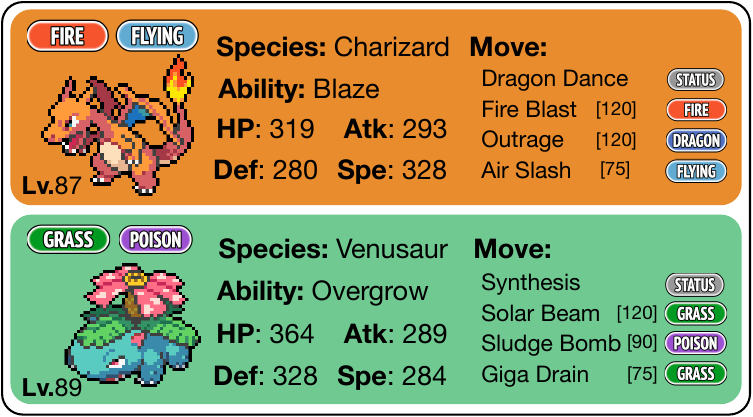}
    \caption{Two representative Pokémon: \textit{Charizard} and \textit{Venusaur}. Each Pokémon has type(s), ability, stats and four battle moves.}
    \vspace{0.4cm}
    \label{fig:pokemon_example}
\end{figure}

\textbf{Species:} There are more than 1,000 Pokémon species~\cite{bulbapedia2024pokemon}, each with its unique ability, type(s), statistics (stats) and battle moves. \textbf{Figure}~\ref{fig:pokemon_example} shows two representative Pokémon: \textit{Charizard} and \textit{Venusaur}.

\begin{figure}[tbp]
    \centering
    \includegraphics[width=5.3cm]{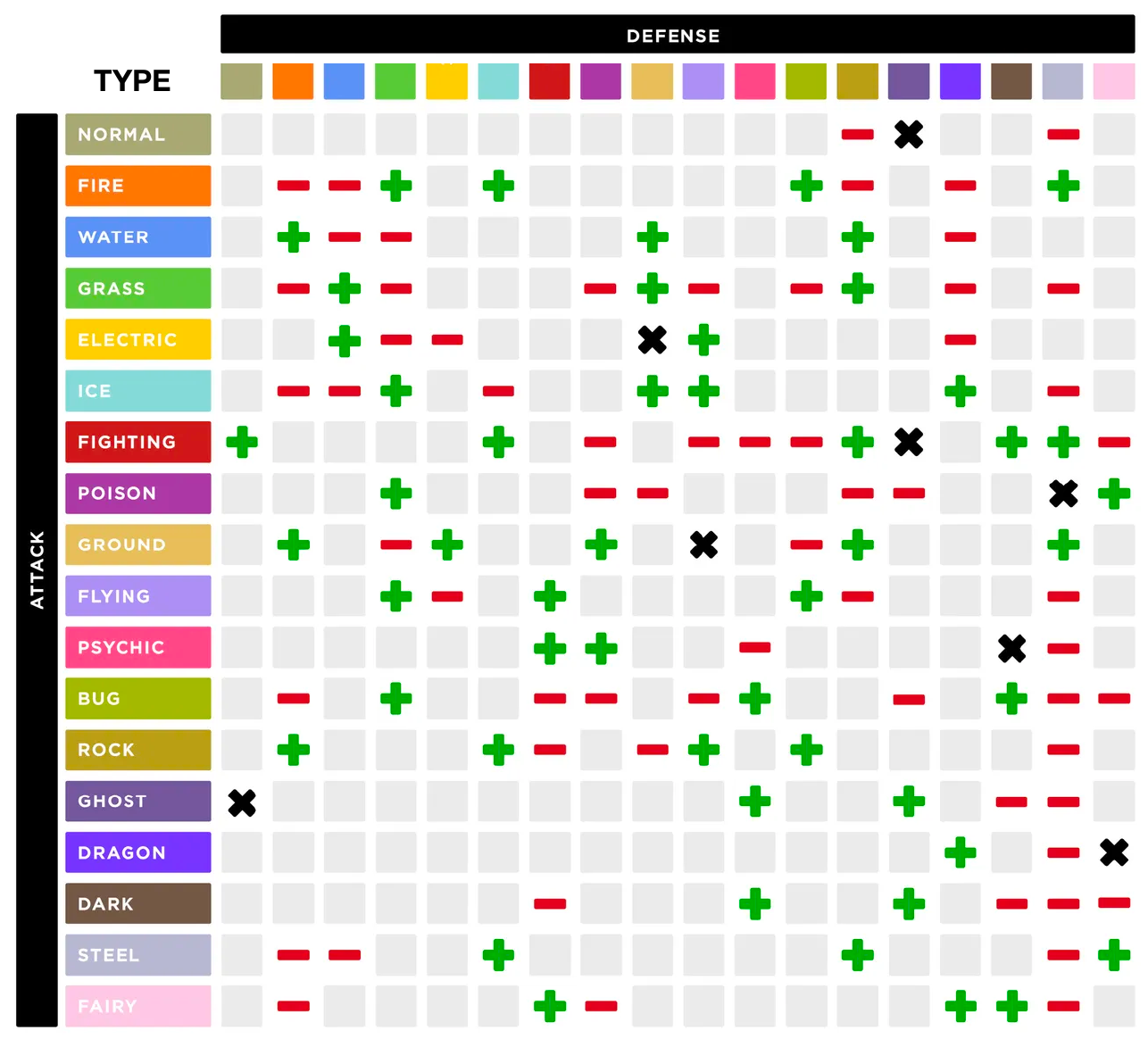}
    \vspace{-0.2cm}
    \caption{Type advantage/weakness relationship. ``$+$'' denotes super-effective (2x damage); ``$-$'' denotes ineffective (0.5x damage); ``$\times$'' denotes no effect (0x damage). Unmarked is standard (1x) damage.}
    \vspace{0.3cm}
    \label{fig:type_advantage}
\end{figure}

\textbf{Type:} Each Pokémon species has up to two elemental types, which determine its advantages and weaknesses. \textbf{Figure}~\ref{fig:type_advantage} shows the advantage/weakness relationship between 18 types of attack moves and attacked Pokémon. For example, fire-type moves like ``Fire Blast" of \textit{Charizard} can cause double damage to grass-type Pokémon like \textit{Venusaur}, while \textit{Charizard} is vulnerable to water-type moves. 

\textbf{Stats:} Stats determine how well a Pokémon performs in battles. There are four stats: (1) Hit Points (HP): determines the damage a Pokémon can take before fainting; (2) Attack (Atk): affects the strength of attack moves; (3) Defense (Def): dictates resistance against attacks; (4) Speed (Spe): determines the order of moves in battle. 


\textbf{Ability:} Abilities are passive effects that can affect battles. For example, \textit{Charizard}'s ability is ``Blaze'', which enhances the power of its fire-type moves when its HP is low. 


\textbf{Move:} A Pokémon can learn four battle moves, categorized as attack moves or status moves. An attack move deals instant damage with a power value and accuracy, and associated with a specific type, which often correlates with the Pokémon's type but does not necessarily align;  A status move does not cause instant damage but affects the battle in various ways, such as altering stats, healing or protect Pokémon, or battle conditions, \textit{etc}. There are 919 moves in total with distinctive effect~\cite{bulbapedia2024listofmoves}. 



\subsection{Battle Rule}
\label{sec:battle_rule}

In one-to-one random battles~\cite{GameplayOfPokemon2023}, two battlers face off, each with six randomly selected Pokémon. Initially, each battler sends out one Pokémon onto the field, keeping the others in reserve for future switches. The objective is to make all the opponent’s Pokémon faint (by reducing their HP to zero) while ensuring that at least one of own Pokémon remains unfainted. The battle is turn-based: at the start of each turn, both players choose an action to perform. Actions fall into two categories: (1) taking a move, or (2) switching to another Pokémon. The battle engine executes actions and updates the battle state for the next step. If a Pokémon faints after a turn and the battler has other Pokémon unfainted, the battle engine forces a switch, which does not count as the player’s action for the next step. After a forced switch, the player can still choose a move or make another switch.





\section{Battle Environment}

\begin{figure*}[htbp]
    \centering
    \includegraphics[width=15.5cm]{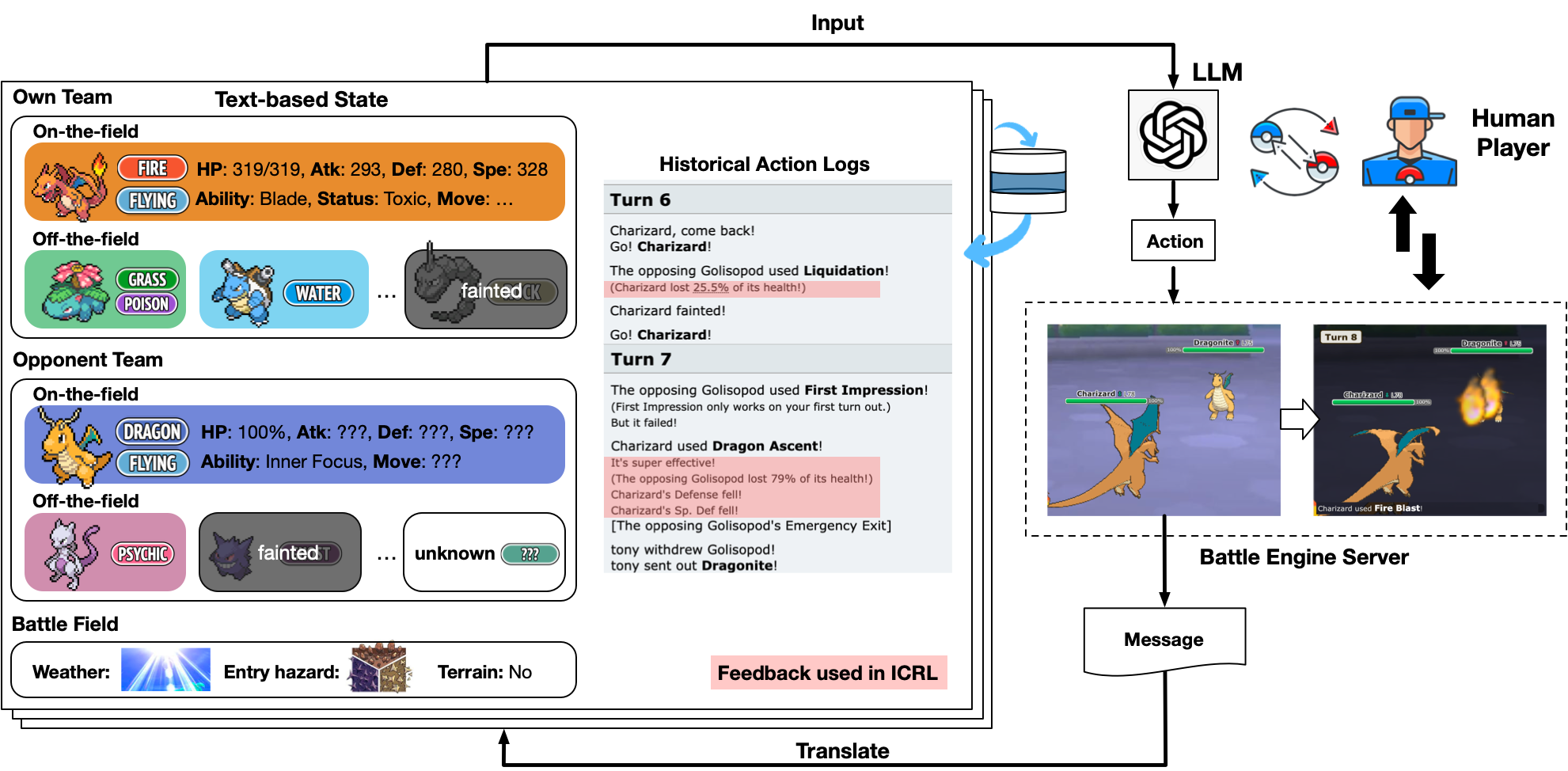}
    \vspace{-0.2cm}
    \caption{The framework that enables LLMs to battle with human players: It parses the messages received from the battle server and translates state logs into text. LLMs take these state descriptions and historical turn logs as input and generates an action for the next step. The action is then sent to the battle server and executed alongside the action chosen by the opponent player.}
    \label{fig:battle_system}
\end{figure*}

\textbf{Battle Engine:} The environment interacts with a battle engine server called Pokémon showdown~\cite{pokemonshowdown2024}, which provides a web-based GUI for human players, as well as web APIs for interacting with message in defined formats.

\textbf{Battle Environment:} We implement a battle environment based on~\cite{poke_env} to support LLMs autonomously play Pokémon battles. \textbf{Figure}~\ref{fig:battle_system} illustrates how the entire framework works. At the beginning of a turn, the environment get an action-request message from the server, including the execution result from the last turn. The environment first parses the message and update local state variables, and then translates the state variables into text. The text description primarily consists of four parts: (1) Own team information, including the attributes of Pokémon both on-the-field and off-the-field; (2) Opponent team information including the attributes of opposing Pokémon on-the-field and off-the-field (some are unknown); (3) Battle field information like the weather, entry hazard and terrain; (4) Historical turn log information, including previous actions of both side Pokémon, which is stored in a log queue. LLMs take the translated state as input and output an action for the next step. The action is sent to the server and executed alongside the action chosen by the human player.


\section{Preliminary Evaluation}

To gain insights into the challenges associated with Pokémon battles, we evaluate the abilities of existing LLMs, including GPT-3.5~\cite{chatGPT}, GPT-4~\cite{GPT4}, and LLaMA-2~\cite{llama2}, 

\subsection{Pokémon Battles}

Placing LLMs in direct competitions against human players is time-consuming as human needs time to think (4 minutes for 1 battle in average). To save time, we adopt a heuristic bot~\cite{pokeenvBaselines2024} to initially battle against human players in the Ladder competitions, and then use the bot to benchmark existing LLMs. The bot is programmed to use status boosting moves, set entry hazards, selecting the most effective actions by considering the stats of Pokémon, the power of moves, and type advantages/weaknesses.

\begin{table}[htbp]
\small
\centering
\caption{\small{Performance of LLMs in battles against the bot.}}
\vspace{0.1cm}
\setlength{\tabcolsep}{2.0mm}
\begin{tabular}{l|cccc}
\toprule
\textbf{Player} & \textbf{Win rate $\uparrow$} & \textbf{Score $\uparrow$} &\textbf{Turn \#} &\textbf{Battle \#}\\ 
\midrule
Human & 59.84\% & 6.75 & 18.74 &  254 \\
Random & 1.2\% & 2.34 & 22.37 & 200 \\
MaxPower & 10.40\% & 3.79 & 18.11 & 200 \\
LLaMA-2 & 8.00\%  & 3.47  & 20.98 &  200\\
GPT-3.5 &  4.00\%  & 2.61  & 20.09 & 100\\
GPT-4  &  26.00\%  & 4.65  & 19.46 & 100\\
\bottomrule         
\end{tabular}
\label{tab:preliminary_result}
\vspace{0.3cm}
\end{table}

The statistic results are presented in \textbf{Table}~\ref{tab:preliminary_result}, where the battle score is defined as the sum of the numbers of the opponent's fainted Pokémon and the player's unfainted Pokémon at the end of a battle. Consequently, the opponent player's battle score is equal to 12 minus the player's battle score. Random is a simple strategy that randomly generates an action every time, and MaxPower chooses the move with the highest power value. Obviously, GPT-3.5 and LLaMA-2 are just slightly better than Random and even GPT-4 cannot beat the bot, let along well-disciplined human players from the Ladder competitions.

By observing LLMs play battles and analyzing the explanations generated with their actions, we identify the occurrence of hallucination~\cite{rawte2023survey,cabello2023pokemonchat}: LLMs can mistakenly claim non-existent type-advantage relationships or, even worse, reverse the advantage relationships between types like sending a grass-type Pokémon to face with a fire-type Pokémon. A clear understanding of type advantage/weakness is crucial in Pokémon battles, as choosing a Pokémon with a type advantage can result in dealing more damage and sustaining less.

\subsection{Test of Hallucination}

To assess hallucination in the outputs of LLMs, we construct the task of type advantage/weakness prediction. The task involves asking LLMs to determine if an attack of a certain type is A. super-effective (2x damage), B. standard (1x damage), C. ineffective (0.5x damage) or D. no effect (0x damage) against a certain type of Pokémon. The 324 (18x18) testing pairs are constructed based on \textbf{Figure}~\ref{fig:type_advantage}.

\begin{table}[htbp]
\centering
\caption{Confusion matrices for type advantage prediction.}
\vspace{0.1cm}
\scalebox{0.75}{
\begin{tabular}{c|cccc|cccc|cccc}
\toprule
\multicolumn{1}{l|}{\textbf{Model}} & \multicolumn{4}{c|}{\textbf{LLaMA-2}}                & \multicolumn{4}{c|}{\textbf{GPT-3.5}}             & \multicolumn{4}{c}{\textbf{GPT-4}}                \\ \midrule
Class & A & B & C & D & A & B & C & D & A & B & C & D \\ \midrule
A & \cellcolor{gray!50}5 & 46 & 0 & 0 & \cellcolor{gray!50}0 & 0 & 49 & 2 & \cellcolor{gray!50}37 & 8 & 5 & 1 \\
B & 25 & \cellcolor{gray!50}179 & 0 & 0 & 2 & \cellcolor{gray!50}6 & 185 & 11 & 0 & \cellcolor{gray!50}185 & 17 & 2 \\
C & 15 & 46 & \cellcolor{gray!50}0 & 0 & 0 & 2 & \cellcolor{gray!50}57 & 2 & 3 & 24 & \cellcolor{gray!50}32 & 2 \\
D & 1 & 7 & 0 & \cellcolor{gray!50}0 & 0 & 0 & 7 & \cellcolor{gray!50}1 & 0 & 0 & 0 & \cellcolor{gray!50}8 \\ \bottomrule
\end{tabular}
\label{tab:confusion_matrix}}
\vspace{0.2cm}
\end{table}

\textbf{Table}~\ref{tab:confusion_matrix} shows the three confusion matrices of LLMs, where their performance is highly related to their win rates in \textbf{Table}~\ref{tab:preliminary_result}. LLaMA-2 and GPT-3.5 suffer from severe hallucination problems, while GPT-4 achieves the best performance with an accuracy of 84.0\%, we still observe it frequently making ineffective actions, which is because in a single battle, LLMs need to compare the types of all the opponent's Pokémon with types of all their Pokémon, as well as types of moves.

\section{\textsc{PokéLLMon}}

\begin{figure}[tbp]
    \centering
    \includegraphics[width=8.7cm]{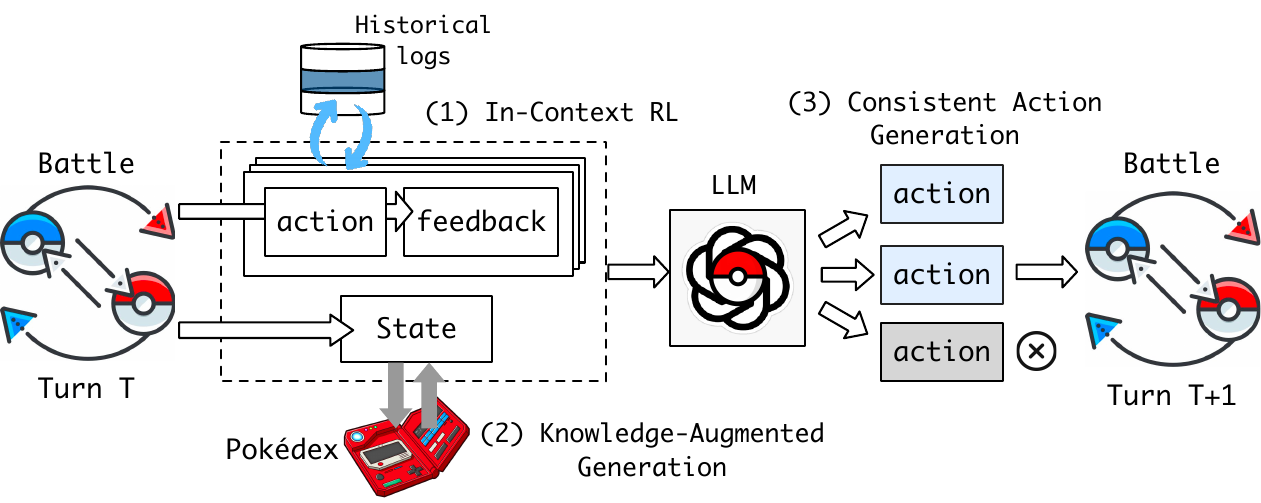}
    \vspace{-0.5cm}
    \caption{{\small \textsc{PokéLLMon}} is equipped with three strategies: (1) ICRL that leverages instant feedbacks from the battle to iteratively refine generation; (2) KAG that retrieves external knowledge to combat hallucination and to act timely and properly; (3) Consistent Action Generation to prevent the panic switching problem.}
    \label{fig:PokeLLMon}
    \vspace{0.4cm}
\end{figure}

\textbf{Overview:} The overall framework of {\small \textsc{PokéLLMon}} is illustrated in \textbf{Figure}~\ref{fig:PokeLLMon}. In each turn, {\small \textsc{PokéLLMon}} uses previous actions and corresponding text-based feedback to \textit{iteratively} refine the policy, and also augments the current state information with external knowledge, such as type advantage/weakness relationships and move/ability effects. Given above information as input, it independently generates multiple actions and selects the most consistent ones as the final output for execution. 


\subsection{In-Context Reinforcement Learning (ICRL)}

\begin{figure}[tbp]
    \centering
    \includegraphics[width=8.0cm]{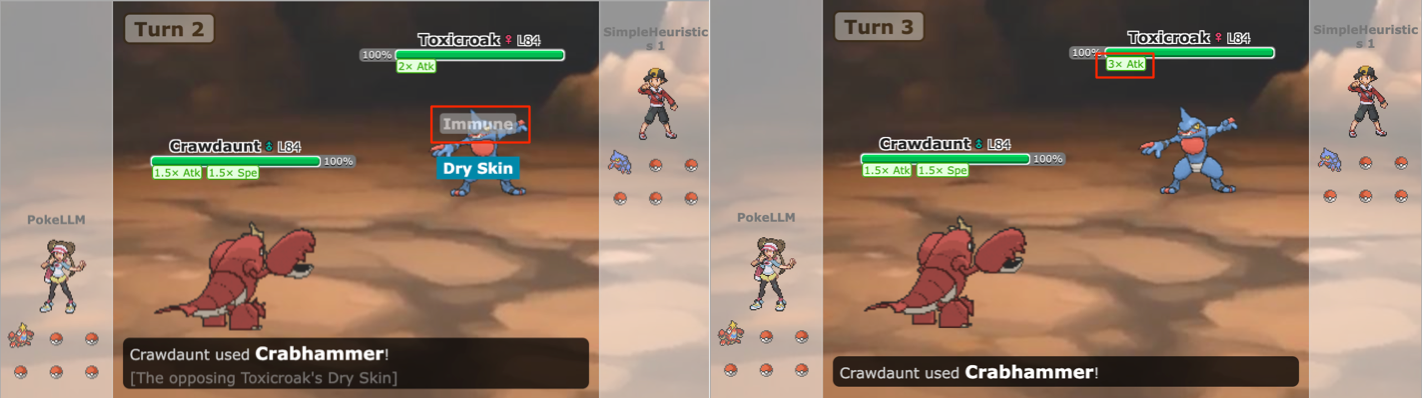}
    \vspace{-0.3cm}
    \caption{The agent repeatedly uses the same attack move but has zero effect to the opposing Pokémon due to its ability ``Dry Skin.'' }
    \vspace{0.2cm}
    \label{fig:no_feedback_case}
\end{figure}

Human players make decisions based not only on the current state but also on the (implicit) feedback from previous actions, such as the change in a Pokémon's HP over two consecutive turns following an attack by a move. Without feedback provided, the agent can continuously stick to the same erroneous action. As illustrated in \textbf{Figure}~\ref{fig:no_feedback_case}, the agent uses ``Crabhammer'', a water-type attack move against the opposing \textit{Toxicroak}, a Pokémon with the ability ``Dry Skin'', which can nullify damage from water-type moves. The ``Immune'' message displayed in the battle animation can prompt a human player to change actions even without knowledge of ``Dry Skin'', however, is not included in the state description. As a result, the agent repeats the same action, inadvertently giving the opponent two free turns to triple \textit{Toxicroak}'s attack stats, leading to defeat.

Reinforcement Learning~\cite{ppo,policygradient,hafner2023mastering} requires numeric rewards to evaluate actions for refining policy. As LLMs can understand languages and distinguish what is good and bad, text-based feedback description provides a new form of ``reward''. By incorporating text-based feedback from the previous turns into the context, the agent is able to refine its ``policy'' \textit{iteratively} and \textit{instantly} during serving, namely In-Context Reinforcement Learning (ICRL).

In practice, we generate four types of feedback: (1) The change in HP over two consecutive turns, which reflects the actual damage caused by an attack move; (2) The effectiveness of attack moves, indicating whether they are super-effective, ineffective, or have no effect (immunity) due to type advantages or ability/move effects; (3) The priority of move execution, providing a rough estimate of speed, as precise stats for the opposing Pokémon are unavailable; (4) The actual effects of executed moves: both status and certain attack moves can cause outcomes like stat boosts or debuffs, recover HP, inflict conditions such as poison, burns, or freezing, \textit{etc}. \textbf{Figure}~\ref{fig:battle_system} presents several instances of generated text-based feedback for ICLR.

\begin{figure}[tbp]
    \centering
    \includegraphics[width=8.0cm]{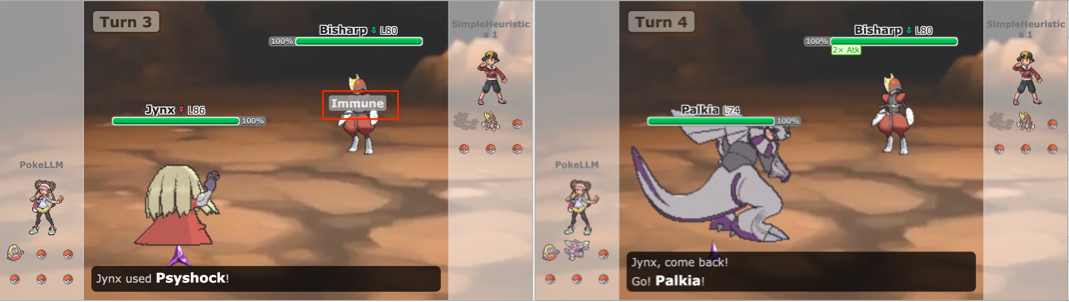}
    \vspace{-0.3cm}
    \caption{In turn 3, the agent uses ``Psyshock'', which cause zero damage to the opposing Pokémon. With ICRL, the agent switch to another Pokémon.}
    \vspace{0.4cm}
    \label{fig:icrl_case}
\end{figure}

\begin{table}[htbp]
\centering
\footnotesize
\caption{\small{Performance of ICRL in battles against the bot.}}
\setlength{\tabcolsep}{2.0mm}
\begin{tabular}{l|cccc}
\toprule
\textbf{Player} & \textbf{Win rate $\uparrow$} & \textbf{Score $\uparrow$} &\textbf{Turn \#} &\textbf{Battle \#}\\ 
\midrule
Human & 59.84\% & 6.75 & 18.74 &  254 \\
Origin &  26.00\%  & 4.65  & 19.46 & 100 \\
ICRL  & 36.00\% & 5.25 & 20.64 & 100 \\
\bottomrule         
\end{tabular}
\label{tab:ICRL_result}
\vspace{0.2cm}
\end{table}

\textbf{Table}~\ref{tab:ICRL_result} shows the improvement brought by ICRL. Compared to the original performance of GPT-4, the win rate is boosted by 10\%, and the battle score increases by 12.9\%. During the battles, we observe that the agent begins to change its action if the moves in previous turns do not meet the expectation, as shown in \textbf{Figure}~\ref{fig:icrl_case}: After observing that the opposing Pokémon is immune to the attack, it switches to another Pokémon.

\subsection{Knowledge-Augmented Generation (KAG)}

Although ICRL can mitigate the impact of hallucination, it can still cause fatal consequences before the feedback is received. For example, if the agent sends out a grass-type Pokémon against a fire-type Pokémon, the former is likely be defeated in a single turn before the agent realize it is a bad decision. To further reduce hallucination, Retrieval-Augmented Generation~\cite{lewis2020retrieval,guu2020retrieval,patil2023gorilla} employ external knowledge to augment generation. In this section, we introduce two types of external knowledge to fundamentally mitigate hallucination.

\textbf{Type advantage/weakness relationship:} In the original state description in \textbf{Figure}~\ref{fig:battle_system}, we annotate all the type information of Pokémon and moves to let the agent infer the type advantage relationship by itself. To reduce the hallucination contained in the reasoning, we explicitly annotate the type advantage and weakness of the opposing Pokémon and our Pokémon with descriptions like ``\textit{Charizard} is strong against grass-type Pokémon yet weak to the fire-type moves''.

\begin{table}[tbp]
\small
\centering
\vspace{-0.2cm}
\caption{\small{Performance of KAG in battles against the bot.}}
\setlength{\tabcolsep}{2.0mm}
\begin{tabular}{l|cccc}
\toprule
\textbf{Player} & \textbf{Win rate $\uparrow$} & \textbf{Score $\uparrow$} &\textbf{Turn \#} &\textbf{Battle \#} \\ 
\midrule
Human & 59.84\% & 6.75 & 18.74 &  254 \\
Origin  & 36.00\% & 5.25 & 20.64 & 100 \\
KAG[Type]  & 55.00\% & 6.09 & 19.28 & 100 \\
KAG[Effect]  & 40.00\% & 5.64 & 20.73 & 100 \\
KAG  & 58.00\% & 6.53 & 18.84  & 100 \\
\bottomrule         
\end{tabular}
\label{tab:kag_result}
\vspace{0.3cm}
\end{table}

\begin{figure}[tbp]
    \centering
    \includegraphics[width=8.0cm]{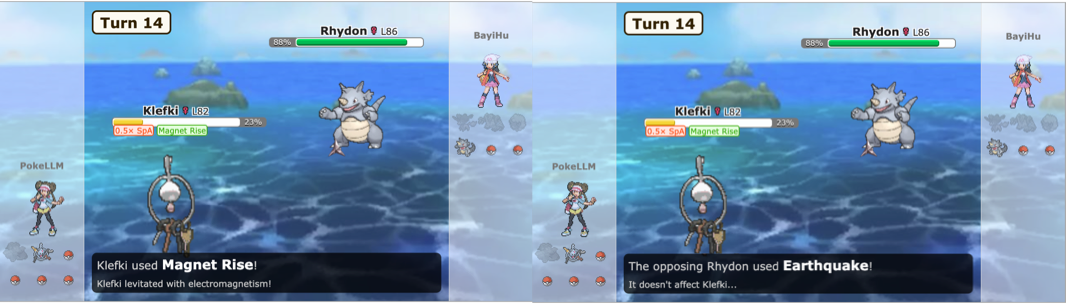}
    \vspace{-0.3cm}
    \caption{The agent understands the move effect and uses it properly: \textit{Klefki} is vulnerable to the ground-type attack of \textit{Rhydon}. Instead of switching, the agent uses ``Magnet Rise'', a move that protects itself from the ground-type attack for five turns, invalidating the ground-type attack ``Earthquake'' of the opposing \textit{Rhydon}.}
    \vspace{0.3cm}
    \label{fig:move_effect_example}
\end{figure}

\textbf{Move/ability effect:} Given the numerous moves and abilities with distinct effects, it is challenging to memorize all of them even for experienced human players. For instance, it's difficult to infer the effect of a status move based solely on its name: ``Dragon Dance'' can boost the user's attack and speed by one stage, whereas ``Haze'' can reset the boosted stats of both Pokémon and remove abnormal statuses like being burnt. Even attack moves can have additional effects besides dealing damage. 

We collect all the effect descriptions of moves, abilities from Bulbapedia~\cite{bulbapedia2024listofmoves,bulbapedia2024abilities} and store them into a Pokédex, an encyclopaedia in Pokémon games. For each Pokémon on the battlefield, its ability effect and move effects are retrieved from the Pokédex and added to the state description.


\textbf{Table}~\ref{tab:kag_result} shows the results of generations augmented with two types of knowledge, where type advantage relationship (KAG[Type]) significantly boosts the win rate from 36\% to 55\%, whereas, Move/ability effect descriptions also enhance the win rate by 4 AP. By combining two of them, KAG achieves a win rate of 58\% against the heuristic bot, approaching a level competitive with human.


With external knowledge, we observe that the agent starts to use very special moves at proper time. As an example shown in \textbf{Figure}~\ref{fig:move_effect_example}, a steel-type \textit{Klefki} is vulnerable to the ground-type attack of the opposing \textit{Rhydon}, a ground-type Pokémon. Usually in such a disadvantage, the agent will choose to switch to another Pokémon, however, it chooses to use the move ``Magnet Rise'', which levitates the user to make it immune to ground-type moves for five turns. As a result, the ground-type attack ``Earthquake'' of the opposing \textit{Rhydon} becomes invalid.



\subsection{Consistent Action Generation}

Existing studies~\cite{CoT,yao2022react,shinn2023reflexion,bommasani2021opportunities,hu2023large} show that reasoning and prompting can improve the ability of LLMs on solving complex tasks. Instead of generating a one-shot action, we evaluate existing prompting approaches including Chain-of-Thought~\cite{CoT} (CoT), Self-Consistency~\cite{SC} (SC) and Tree-of-Thought~\cite{ToT} (ToT). For CoT, the agent initially generates a thought that analyzes the current battle situation and outputs an action conditioned on the thought. For SC (k=3), the agent generates three times of actions and select the most voted answer as the output. For ToT (k=3), the agent generates three action options and picks out the best one evaluated by itself.

\begin{table}[htbp]
\small
\centering
\caption{\small{Performance of prompting approaches in battles against the bot.}}
\setlength{\tabcolsep}{2.0mm}
\begin{tabular}{l|cccccc}
\toprule
\textbf{Player} & \textbf{Win rate $\uparrow$} & \textbf{Score $\uparrow$} & \textbf{Turn \#} &\textbf{Battle \#} \\ 
\midrule
Human & 59.84\% & 6.75 & 18.74 &  254 \\
Origin & 58.00\% & 6.53 & 18.84  & 100 \\
CoT  & 54.00\% & 5.78  & 19.60 & 100 \\
SC (k=3)   & \textbf{64.00}\% & 6.63 & 18.86 & 100 \\
ToT (k=3)  & 60.00\% & 6.42 & 20.24 & 100 \\
\bottomrule         
\end{tabular}
\label{tab:reasoning}
\vspace{0.3cm}
\end{table}

\textbf{Table}~\ref{tab:reasoning} presents the comparison results of the original IO prompt generation and three algorithms. Notably, CoT results in a performance degradation by a 6 AP drop in the win rate. In comparison, SC brings a performance improvement, with the win rate surpassing human players. Beyond the results, our greater interest lies in understanding the underlying reasons for these observations.


\begin{figure}[tbp]
    \centering
    \includegraphics[width=8.0cm]{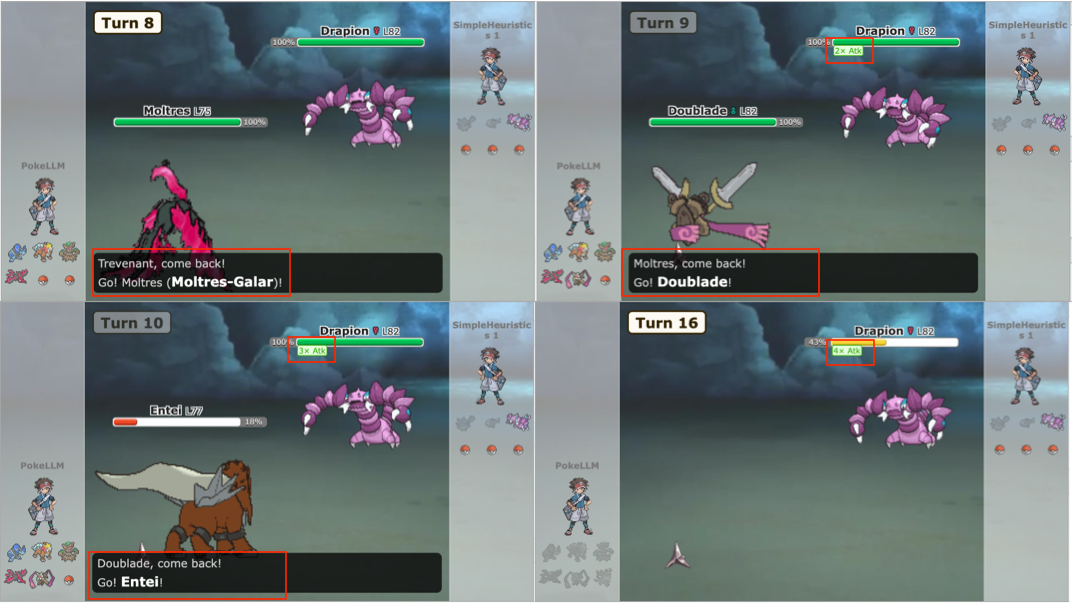}
    \vspace{-0.3cm}
    \caption{When facing a powerful Pokémon, the agent with CoT switches different Pokémon in three consecutive to elude the battle. This gives the opponent three free turns to quadruple its attack stats and quickly defeat the agent's entire team.}
    \vspace{0.3cm}
    \label{fig:panic_switch}
\end{figure}

As introduced in Section~\ref{sec:battle_rule}, for each turn there is single action can be taken, which means if the agent chooses to switch yet the opponent choose to attack, the switch-in Pokémon will sustain the damage. Usually switching happens when the agent decides to leverage the type advantage of an off-the-battle Pokémon, and thus the damage taken is sustainable since the switch-in Pokémon is typically type-resistant to the opposing Pokémon's moves. However, when the agent with CoT reasoning faces a powerful opposing Pokémon, its actions become inconsistent by switching to different Pokémon in consecutive turns, which we call \textit{panic switching}. \textit{Panic switching} wastes chances of taking moves and leading to the defeat. An illustrative example is shown in \textbf{Figure}~\ref{fig:panic_switch}: starting from turn 8, the agent chooses to continuously switch to different Pokémon in three consecutive turns, giving the opposing Pokémon three free turns to boost its attack stats to four times and take down the agent's entire team quickly.


\begin{table}[htbp]
\footnotesize
\centering
\caption{\small{Statistic analysis of panic switching}}
\setlength{\tabcolsep}{1.4mm}
\begin{tabular}{l|ccccc}
\toprule
\textbf{Player} & \textbf{Win rate $\uparrow$} & \textbf{Switch rate} & \textbf{CS1 rate} & \textbf{CS2 rate} \\ 
\midrule
Origin & 58.00\%  & 17.05\% & 6.21\% & 22.98\% \\
CoT  & 54.00\%  & 26.15\% & 10.77\%  & 34.23\%   \\
SC (k=3) & 64.00\% & 16.00\% & 1.99\% & 19.86\% \\
ToT (k=3) & 60.00\% & 19.70\% & 5.88\% & 23.08\% \\
\bottomrule         
\end{tabular}
\label{tab:panic_switch_static}
\vspace{0.3cm}
\end{table}

\textbf{Table}~\ref{tab:panic_switch_static} provides statistical evidence, where CS1 represents the ratio of active switches where the last-turn action is a switch and CS2 rates represent the ratio of active switches here at least one action from the last two turns is a switch, among all active switches, respectively. The higher the CS1 rate, the greater the inconsistency of generation. Obviously, CoT largely increases the continuous switch rate, whereas, SC decreases the continuous switch rate.

Upon examining the thoughts generated by CoT, we observe that the thoughts contain panic feelings: the agent describes how powerful the opposing Pokémon is and the weaknesses of the current Pokémon, and ultimately decides to switch to another Pokémon, as in ``\textit{Drapion} has boosted its attack to two times, posing a significant threat that could potentially knock out \textit{Doublade} with a single hit. Since \textit{Doublade} is slower and likely to be knocked out, I need to switch to \textit{Entei} because...''. Action generation conditioned on panic thoughts leads the agent to continuously switch Pokémon instead of attacking. In comparison, consistent action generation with SC decreases the continuous switch ratio by independently generating actions multiple times and voting out the most consistent action as shown in \textbf{Figure}~\ref{fig:PokeLLMon}, leading to a higher win rate. The observation is reflecting: when humans face stressful situations, overthinking and exaggerating difficulties lead to panic feelings and paralyze their ability to take actions, leading to even worse situations.

\section{Online Battle}

To test the battle ability of {\small \textsc{PokéLLMon}} against human, we set up the eighth-gen battles on Pokémon Showdown, where the agent battled against random human players for the Ladder competitions from Jan. 25 to Jan. 26, 2024. Besides, we invited an human player who has over 15 years of experience with Pokémon games, representing the average ability of human players to play against {\small \textsc{PokéLLMon}}. 


\subsection{Battle Against Human Players}


\begin{table}[htbp]
\small
\centering
\caption{\small{Performance of {\small \textsc{PokéLLMon}} against human players.}}
\setlength{\tabcolsep}{1.8mm}
\begin{tabular}{l|cccccc}
\toprule
\textbf{v.s. Player} & \textbf{Win rate $\uparrow$} & \textbf{Score $\uparrow$} &\textbf{Turn \#} &\textbf{Battle \#} \\ 
\midrule
Ladder Player & 48.57\% & 5.76 & 18.68 & 105 \\
Invited Player & 56.00\% & 6.52 & 22.42 & 50 \\
\bottomrule         
\end{tabular}
\label{tab:vs_human}
\vspace{0.3cm}
\end{table}

\textbf{Table}~\ref{tab:vs_human} presents the performance of the agent against human players. {\small \textsc{PokéLLMon}} demonstrates comparable performance to disciplined Ladder players who have extensive battle experience, and achieves a higher win rate than the invited player. The average number of turns in the Ladder competitions is lower because human players sometimes forfeit when they believe they will lose to save time.

\subsection{Battle Skill Analysis}

\begin{figure}[tbp]
    \centering
    \includegraphics[width=8.0cm]{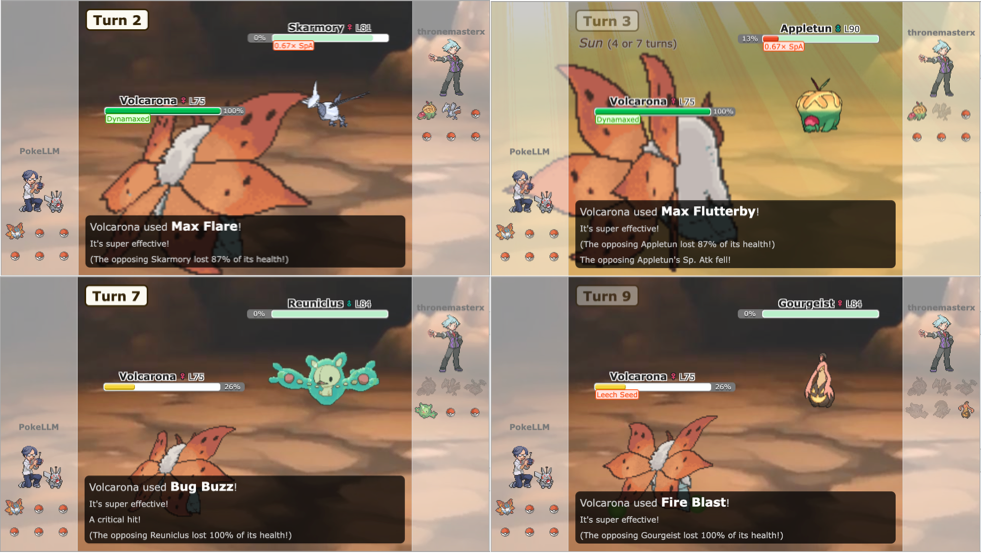}
    \vspace{-0.3cm}
    \caption{{\small \textsc{PokéLLMon}} selects effective moves in every turn, causing the opponent's entire team to faint using one Pokémon.}
    \vspace{0.3cm}
    \label{fig:choose_action_example}
\end{figure}

\textbf{Strength:} {\small\textsc{PokéLLMon}} seldom make mistakes at choosing the effective move and switching to another suitable Pokémon due to the KAG strategy. As shown in \textbf{Figure}~\ref{fig:choose_action_example}, in one battle, the agent uses only one Pokémon to cause the entire opponent team fainted by choosing different attack moves toward different Pokémon.

Moreover, {\small\textsc{PokéLLMon}} exhibits human-like attrition strategy: With some Pokémon have the ``Toxic'' move that can inflict additional damage every turn and the ``Recover'' move that can recover its HP, the agent starts to first poisoned the opposing Pokémon and frequently uses the ``Recover'' to prevent itself from fainting. By prolonging the battle, the opposing Pokémon's HP is gradually depleted by the poisoning damage. Using attrition strategy requires an understanding of moves like ``Toxic'', ``Recover'' and ``Protect'', as well as the right timing for their use (such as when there's no type-weakness or when having high defense). An example with battle animation can be found at: {\small \url{https://poke-llm-on.github.io}}.

\begin{figure}[tbp]
    \centering
    \includegraphics[width=8.0cm]{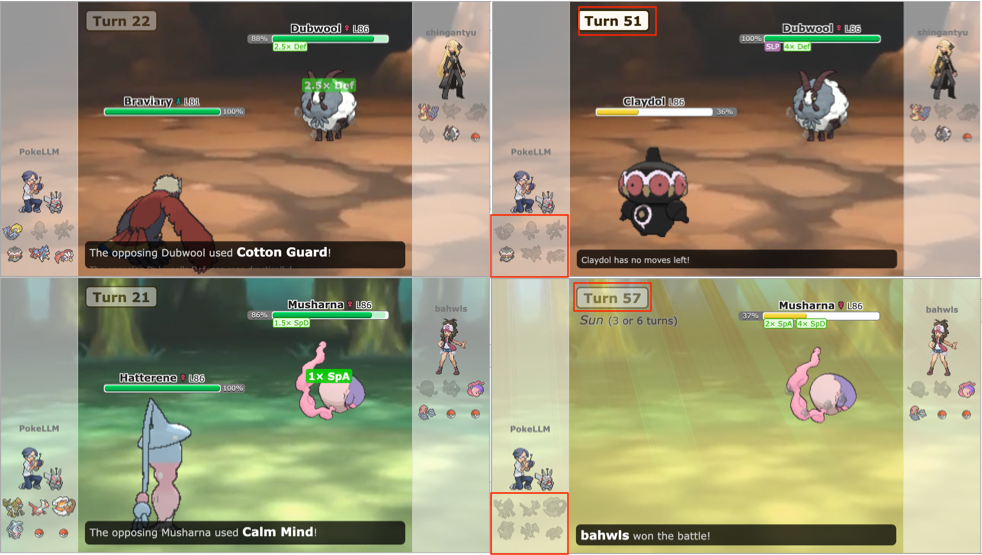}
    \vspace{-0.3cm}
    \caption{{\small\textsc{PokéLLMon}} suffers from attrition strategies: the opponent players frequently recover high-defense Pokémons. Breaking the dilemma requires joint effects across many turns.}
    \vspace{0.4cm}
    \label{fig:attrition_strategy}
\end{figure}

\textbf{Weakness:} {\small\textsc{PokéLLMon}} tends to take actions that can achieve short-term benefits, therefore, making it vulnerable to human players' attrition strategy that requires long-term effort to break. 
As shown in the two battles in \textbf{Figure}~\ref{fig:attrition_strategy}, after many turns, the agent's entire team is defeated by the human players' Pokémon, which have significantly boosted defense and engage in frequent recovery. \textbf{Table}~\ref{tab:attrit} reports the performance of  {\small\textsc{PokéLLMon}} in battles where human players either use the attrition strategy or not. Obviously, in battles without the attrition strategy, it outperforms Ladder players, while losing the majority of battles when human play the attrition strategy.

\begin{table}[htbp]
\small
\centering
\caption{\small{Battle performance impacted by the attrition strategy}}
\setlength{\tabcolsep}{1.8mm}
\begin{tabular}{l|cccc}
\toprule
\textbf{Ladder} & \textbf{Win rate $\uparrow$} &\textbf{Score $\uparrow$} &\textbf{Turn \#} &\textbf{Battle \#} \\ 
\midrule
w. Attrition & 18.75\% & 4.29 & 33.88 & 16 \\
w/o Attrition & 53.93\% & 6.02 & 15.95 & 89     \\
\bottomrule         
\end{tabular}
\label{tab:attrit}
\vspace{0.3cm}
\end{table}

The ``Recover'' move recovers 50\% HP in one turn, which means if an attack cannot cause the opposing Pokémon more than 50\% HP damage in one turn, it will never faint. The key to breaking the dilemma is to firstly boost a Pokémon's attack to a very high stage and then attack to cause unrecoverable damage, which is a long-term goal that requires joint efforts across many turns. {\small\textsc{PokéLLMon}} is weak to the long-term planing because current design does not keep a long-term plan in mind across many timesteps, which will be included in the future work.


\begin{figure}[tbp]
    \centering
    \includegraphics[width=8.0cm]{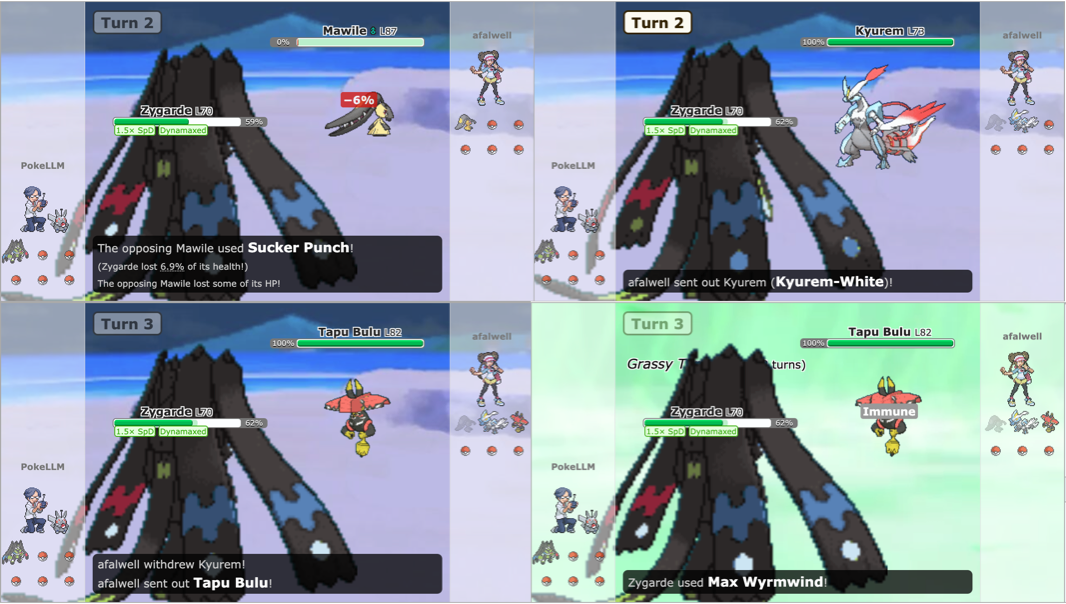}
    \vspace{-0.3cm}
    \caption{An experienced human player misdirects the agent to use a dragon-type attack by firstly sending out a dragon-type Pokémon and immediately switch to another Pokémon immune to the dragon-type attack.}
    \vspace{0.3cm}
    \label{fig:human_trick}
\end{figure}

Finally, we observe that experienced human players can misdirect the agent to bad actions. As shown in \textbf{Figure}~\ref{fig:human_trick}, our \textit{Zygarde} has one chance to use an enhanced attack move. At the end of turn 2, the opposing \textit{Mawile} is fainted, leading to a forced switch and the opponent choose to switch in \textit{Kyurem}. This switch is a trick that lures the agent uses a dragon-type move in turn 3 because \textit{Kyurem} is vulnerable to dragon-type attacks. In turn 3, the opponent switches in \textit{Tapu Bulu} at the beginning, a Pokémon immune to dragon-type attacks, making our enhanced attack chance wasted. The agent is fooled because it makes decision only based on the current state information, while experienced players condition on not only the state information, but also the opponent's next action prediction.

Seeing through tricks and predicting the opponent's next action require the agent being disciplined in the real battle environment, which is the future step in our work.

\section{Conclusion}
In this paper, we enable LLMs to autonomously play the well-known Pokémon battles against human. We introduce {\small \textsc{PokéLLMon}}, the first LLM-based agent that achieves human-competent performance in tactical battle games. We introduce three key strategies in the design of {\small \textsc{PokéLLMon}}: (i) In-Context Reinforcement Learning, which consumes the text-based feedback as ``reward'' to \textit{iteratively} refine the action generation policy without training; (ii) Knowledge-Augmented Generation that retrieves external knowledge to combat hallucination and ensures the agent act timely and properly; (iii) Consistent Action Generation that prevents the \textit{panic switching} issue when encountering powerful opponents. The architecture of {\small \textsc{PokéLLMon}} is \textit{general} and can be adapted for the design of LLM-based agents in many other games, addressing the problems of hallucination and action inconsistency.

Online battles show that {\small \textsc{PokéLLMon}} demonstrates human-like battle ability and strategies, achieving 49\% of win rate in the Ladder competitions and 56\% of win rate in the invited battles. Furthermore, we uncover its vulnerabilities to human players' attrition strategies and deception tricks, which are considered as our future work.



\bibliography{example_paper}
\bibliographystyle{icml2024}


\end{document}